# Novel Topological Machine Learning Methodology for Stream-of-Quality Modeling in Smart Manufacturing

Jay Lee[a], Dai-Yan Ji[a,*], Yuan-Ming Hsu[b]

[a]*Center for Industrial Artificial Intelligence, Department of Mechanical Engineering, University of Maryland, College Park, MD 20742, USA*
[b]*Department of Mechanical & Materials Engineering, College Engineering and Applied Science, University of Cincinnati, OH 45221, USA*

\* Corresponding author. *E-mail address:* jidn@umd.edu (D. Ji)

**Abstract**

This paper presents a topological analytics approach within the 5-level Cyber-Physical Systems (CPS) architecture for the Stream-of-Quality assessment in smart manufacturing. The proposed methodology not only enables real-time quality monitoring and predictive analytics but also discovers the hidden relationships between quality features and process parameters across different manufacturing processes. A case study in additive manufacturing was used to demonstrate the feasibility of the proposed methodology to maintain high product quality and adapt to product quality variations. This paper demonstrates how topological graph visualization can be effectively used for the real-time identification of new representative data through the Stream-of-Quality assessment.

*Keywords:* Topological Data Analysis (TDA), Cyber-Physical Systems (CPS), Stream-of-Quality, Data visualization, Machine learning

## 1. Introduction

In multi-stage manufacturing systems (MMS) [1,2], the task of analyzing operational parameters and sensor data for various stages is a significant challenge due to the complexity of different manufacturing processes. In recent advancements, the Stream-of-Quality (SoQ) [3] framework has offered a methodology for the integrated assessment of product quality indicators across different stages of manufacturing. However, MMS often produce vast, multidimensional datasets that traditional statistical methods struggle to analyze effectively due to the non-linear and dynamic behaviors of these systems. To address these challenges, this paper introduces a topology analytics approach [4,5], emphasizing its advantages in visualizing data shapes, detecting clusters and topological structures, and selecting features for discrimination and interpretability. Most importantly, topology analytics effectively integrate into CPS [6], enabling online monitoring to identify process variability correlations within the SoQ framework. This paper emphasizes the important role of topology analytics in smart manufacturing by improving data analysis and decision-making processes.

## 2. Topological Machine Learning Methodology

### 2.1. Data Selection

Table 1 shows the comparison of original data and representative data. Original data, collected randomly and in high volumes, presents ununiform quality with low representativeness and scarce information density because of redundant or irrelevant information, especially in high-dimensional datasets where the inherent complexity of data space complicates the extraction of significant patterns. In contrast, representative data is selected based on topological criteria, wherein the selection process is guided by the dataset's intrinsic connectivity and continuity properties on topological spaces, enabling the extraction of data that accurately represents the dataset's overall structure. Therefore, the representative data is typically of lower volume but higher quality to possess high representatives due to preserving invariant topological features under continuous transformations. These features represent the richest information to withstand the complexities of high-dimensional datasets.

**Table 1**
The comparison of original data and data with topological representation.

|  | Original Data | Data with topological representation |
|---|---|---|
| Selection | Random | Topology-based selection |
| Data Quantity | High volume | Clustered low volume |
| Data Quality | Ununiform | High |
| Representative | Low representative data | High representative data |
| Informative | Scarce | Rich |

## 2.2. Topological Data Analysis

Topological Data Analysis (TDA) [7,8] transforms complex datasets into simplified but informative representations, using persistent homology (e.g., persistence diagrams and barcodes). Persistence diagrams plot the birth and death of these features, while barcodes represent them as line segments, enhancing the ease of visualization and comparative assessment of topological properties. These TDA's representations, by facilitating the identification of holes or persistent features in the data that remain invariant over a range of scales, provide a powerful way to distinguish patterns that might not be apparent through conventional data analysis techniques.

## 2.3. Adaptive Clustering & Prediction Model

Adaptive clustering and predictive modeling are addressed to enhance analytical ability and predictive accuracy. Adaptive clustering, influenced by the underlying topological structure, can dynamically adjust groups' formation based on shared features. Predictive models perform the development of decision trees, support vector machines, neural networks, and other machine-learning techniques to efficiently evaluate the overall performance gained from TDA to forecast quality trends, optimize production processes, and monitor quality control. Integrating TDA with these predictive models assists on-site engineers in maintaining quality and improving accuracy in decision-making, leading to more informed and effective strategies in smart manufacturing.

## 2.4. Topological Graph Visualization & Update Model

The first phase, employed by the mapper algorithm, analyzes topological graph visualization dynamically to reveal hidden patterns and anomalies, decomposed into three major steps: filtration, which condenses high-dimensional data to preserve critical features; coverage, segmenting the reduced space into overlapping areas to capture diverse data facets; and clustering, utilizing algorithms to group data based on topological proximity. The second phase is to identify new representative data via topological graph visualization to update the model in real-time. These two phases not only enhance the interpretability of complex manufacturing data via topological graph visualization, but also advance manufacturing systems towards an online adaptive, continuously updating model.

## 2.5. Online Analysis

Online analytic frameworks are capable of conducting real-time analysis, including monitoring data distributions across different classes for quickly identifying potential issues, enabling manufacturers to make informed decisions, optimize processes promptly, and proactively adapt to changes in smart manufacturing. Additionally, the real-time analysis of data distribution can also assist in identifying changes in production quality and detecting inefficiencies to move towards more intelligent, data-driven manufacturing processes.

## 3. System Architecture of Implementing Topological Analytics

To implement TDA in smart manufacturing, the 5C level CPS architecture was deployed [6] with compelling case studies. Fig. 1 illustrates how the topological machine learning methodology can be integrated with the 5C level CPS architecture in different applications, such as design & simulation optimization, smart manufacturing, and predictive maintenance. Three principal technologies (i.e., Data Technology (DT), Analytic Technology (AT), and Operation Technology (OT)) can be applied when they are situated in the 5C level CPS architecture (i.e., 5C: Connection, Conversion, Cyber, Cognition, and Configuration).

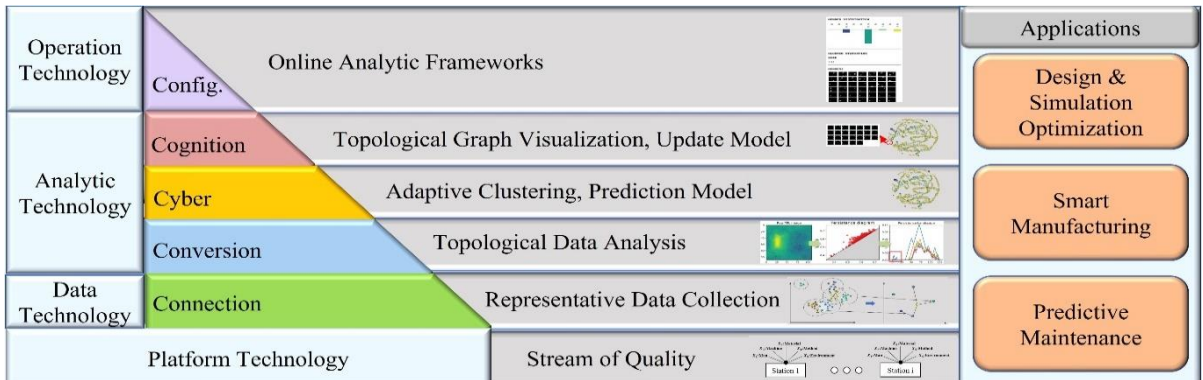

**Fig. 1.** Topological analytics within 5C level CPS architecture and stream-of-quality platform.

## 4. Topological Analytics on SoQ Platform

The flowchart in Fig. 2 illustrates the application of topological analytics for the SoQ (i.e., integration within the platform technology, as shown in Fig.1) assessment of multi-stage manufacturing processes. The process starts with data collection from different manufacturing stations, where data sources, including man, machine, material, method, and environment, are collected to process data preprocessing and topological-based feature extraction, followed by filtering to preserve important features for quality assessment. Online update visualization is initiated to integrate new important features from each new station to generate cluster quality metrics. The final cluster quality represents the stream of quality, encompassing the initial cluster quality, ongoing updates to the cluster quality as the new station is added, and the final cluster quality that feeds back into the process for continuous online updating. When new representative data are identified at the final station based on cluster quality, the online update model is initiated to update labels to the visualization layer and simultaneously output the final prediction quality. Therefore, the total SoQ comprises the process of online updates, outputting content within both the final cluster quality and the final prediction quality.

## 5. Case Study: Topological Analytics for SoQ in Additive Manufacturing System

Two-Photon Lithography (TPL) is an emerging technique in additive manufacturing that fabricates the three-dimensional nanostructure. A significant challenge in TPL is to identify the optimal light dosage parameters, influenced by the scanning speed and laser intensity, during the manufacturing process.

The purpose of this case study, which utilizes data collected [9,10] by Lawrence Livermore National Laboratory (LLNL), implements topological analytics for the SoQ assessment applied to an additive manufacturing system that can online detect the part quality across multiple stages, as shown in Fig. 3. A manufacturing line with eight stages is arranged to process materials into a final product. SoQ can be evaluated by the following outcomes:

I. *Cluster Quality Analysis*: At each stage, the clusters are visualized in a topological graph visualization format, with the node color representing the proportion of every class. While the manufacturing process advances to the next stage, cluster quality is visualized to dynamically update how new data integrates with existing data.

II. *Prediction Quality Analysis*: A predictive quality model, which uses the incoming new data to update its prediction continuously, is designed to forecast the product's quality at subsequent stages, showing the variation of the product's quality from one stage to the final stage, which could be original, uncured, cured, or damaged, as indicated by the color-coded legend.

III. *Integration of New Representative Data*: At the final stage, the model identifies new representative data that significantly differs from the existing clusters. These new representative data, potentially indicating process anomalies or quality deviations, represent a new label for an online update of the model to ensure that the prediction accurately shows the final product's quality and enhances the model's adaptability to process variations.

## 6. Challenges of Topological Analytics in Manufacturing

Employing the mapper algorithm in topological analysis for the SoQ assessment can be visualized to uncover hidden patterns through node property comparisons. Although the proposed framework has proven effective, as shown in section 5: case study, it also highlights key factors for further exploration and challenge.

I. *Parameter Settings*: key considerations include distance metric, filter function, and resolution parameter. In particular, the filter function, which reduces dimensionality (e.g., options like t-SNE, PCA,

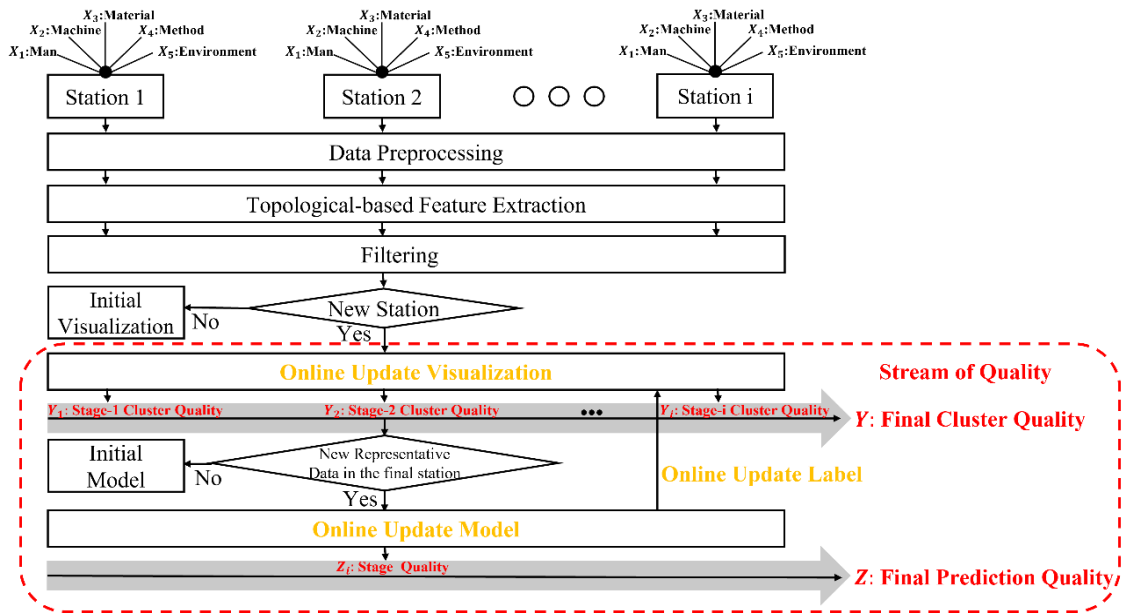

**Fig. 2.** Topological analytics of SoQ on a multi-stage manufacturing system.

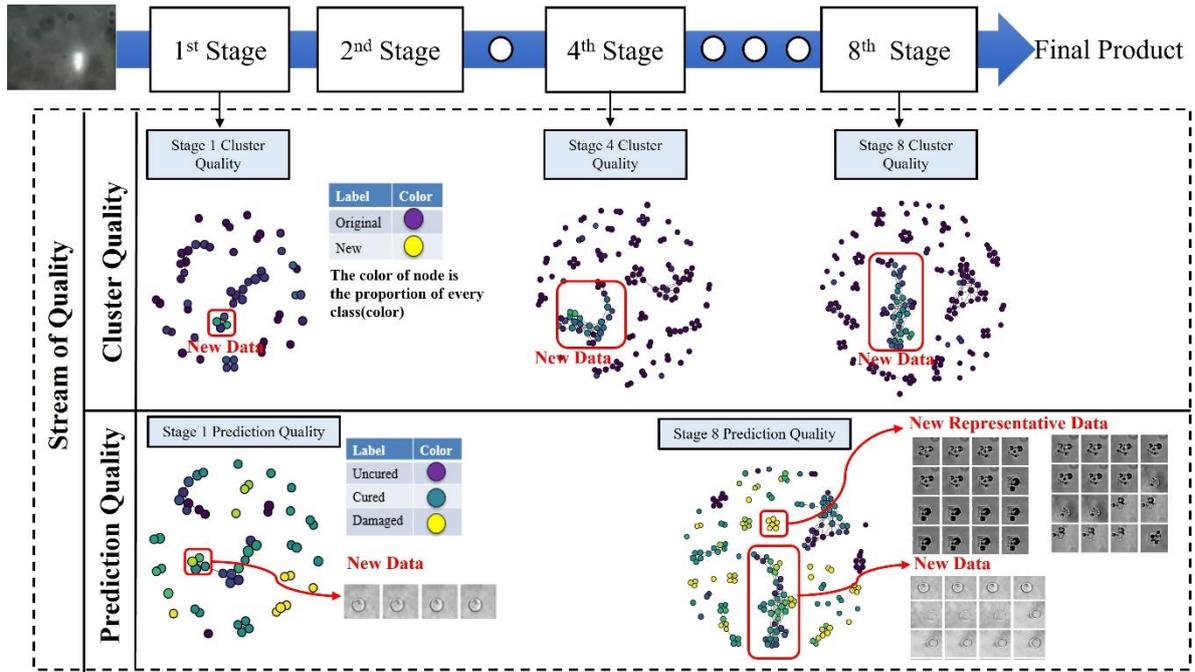

**Fig. 3.** Case study of topological analytics for SoQ applied to Additive Manufacturing System

and UMAP), is crucial for simplifying complex data structures.

II. *Processing Filtered data*: Adjusting two key parameters, both interval and overlap, is essential for analyzing data clusters in each subset, enhancing a comprehensive understanding of data distribution.

## 7. Conclusions

This study emphasizes the importance of employing topological analytics within the SoQ methodology in smart manufacturing. Through a practical case study, the proposed framework demonstrates the critical role of topological analytics in dynamically integrating new representative data and updating the model to refine quality prediction. Its function is to maintain a high-quality final product and adapt the variations of the product quality throughout the manufacturing process. Moreover, SoQ includes both cluster quality and prediction quality to have real-time monitoring, root cause analysis, and the optimization of process-to-process relationships, which ultimately leads to improved product quality.

**Acknowledgments**
This work was performed under the following financial assistance award 60NANB23D228 from U.S. Department of Commerce, National Institute of Standards and Technology.